\documentclass[conference]{IEEEtran}
\IEEEoverridecommandlockouts

\usepackage{cite}
\usepackage{amsmath,amssymb,amsfonts}
\usepackage{algorithmic}
\usepackage{graphicx}
\usepackage{textcomp}
\usepackage{xcolor}
\usepackage{listings}
\usepackage{booktabs}
\usepackage{url}
\usepackage{tabularx}
\usepackage{array}
\usepackage{hyperref}
\lstdefinestyle{pythonstyle}{
    language=Python,
    basicstyle=\small\ttfamily,
    keywordstyle=\color{blue},
    commentstyle=\color{gray},
    stringstyle=\color{orange},
    numbers=none,
    frame=single,
    breaklines=true,
    breakatwhitespace=false,
    columns=flexible,
    keepspaces=true,
    xleftmargin=0pt,
    xrightmargin=0pt
}

\def\BibTeX{{\rm B\kern-.05em{\sc i\kern-.025em b}\kern-.08em
    T\kern-.1667em\lower.7ex\hbox{E}\kern-.125emX}}

\begin{document}

\title{Interpretable Fuzzy Rule-Based Regression Extension for Ex-Fuzzy Library}
\author{\IEEEauthorblockN{Cayan Deniz Kucuktopana, Javier Fumanal-Idocin, Richard Pitts, Javier Andreu-Perez}
\IEEEauthorblockA{\textit{School of Computer Science} \\
\textit{and Electronic Engineering} \\
\textit{University of Essex}\\
Colchester, United Kingdom \\
\{ck23382, j.fumanal-idocin, richard.pitts, j.andreu-perez\}@essex.ac.uk}
}\maketitle

\begin{abstract}
Machine learning models achieve high predictive accuracy in regression tasks, but their deployment in safety-critical and regulated domains requires interpretability. While fuzzy rule-based systems offer transparent, linguistically explicit interpretable models, Mamdani-style fuzzy regression remains underrepresented in modern machine learning software libraries. This paper presents an interpretable regression extension for the Ex-Fuzzy library, enabling Mamdani fuzzy inference with scalar consequents learned directly from data. For this, a target-aware partition initialisation strategy based on Fuzzy C-Means clustering is introduced, in which linguistic variables are derived from an augmented input–output space to emphasise output-relevant regions of the feature space. The proposed extension is evaluated on ten regression datasets from the KEEL repository, comparing Gaussian and trapezoidal partition strategies against standard baselines including linear regression, multilayer perceptrons, and random forests. Experimental results show that Gaussian partitions consistently outperform uniform trapezoidal partitions, achieving a mean coefficient of determination of approximately 0.86 while producing compact rule bases of 10–15 human-readable rules. The proposed implementation provides a transparent and competitive alternative to black-box regression models, supporting practical interpretability with competitive predictive performance.
\end{abstract}

\begin{IEEEkeywords}
Fuzzy systems, Interpretable machine learning, Fuzzy regression, Fuzzy Software,
Genetic algorithms, Mamdani inference, Rule-based models
\end{IEEEkeywords}

\section{Introduction}

Machine learning models increasingly achieve high predictive accuracy across diverse regression tasks, yet their deployment in safety-critical and regulated domains remains challenging due to limited interpretability \cite{rudin2019stop}. Neural networks and ensemble methods, while powerful, operate as black boxes whose decision processes cannot be readily inspected, validated, or modified by domain experts. This opacity creates barriers in applications such as medical diagnosis, financial risk assessment, and environmental monitoring, where understanding why a prediction was made is as important as the prediction itself.

Fuzzy rule-based systems offer a compelling alternative by representing knowledge as linguistically interpretable IF–THEN rules \cite{mendel2017, fumanal2023artxai}. In explainable AI, they are frequently highlighted as a natural route to transparency because they provide structured, human-auditable reasoning. A rule such as ``IF displacement IS High AND weight IS High THEN fuel\_efficiency IS Low'' directly encodes domain relationships in human-readable form, allowing domain experts to inspect, validate, and refine model behaviour using prior knowledge. This interpretability has been successfully leveraged in recent applied work, where genetic optimisation of fuzzy rule structures retained predictive performance while enabling meaningful inspection \cite{andreu2021explainable}.

The Ex-Fuzzy library \cite{exfuzzy2024} provides a Python implementation for constructing and optimising fuzzy rule-based classifiers using genetic algorithms. Its focus on explainability, visualisation, and scikit-learn compatibility has made it accessible to both researchers and practitioners. However, Ex-Fuzzy currently supports only classification tasks, thereby limiting its applicability to many real-world problems that require continuous output prediction. Despite extensive research on fuzzy modelling, Mamdani-style fuzzy regression has received far less attention in the literature than Takagi-Sugeno approaches, as highlighted by recent systematic reviews \cite{chukhrova2019}. While ANFIS \cite{zhang2024scikit, chen2020fuzzyr} have been implemented for regression in Python, R, or MATLAB, at the time of this paper, \emph{there is no Python library that provides a fuzzy-rule system for regression oriented to explainability.} Existing Python libraries such as Simpful provide general-purpose Mamdani/Sugeno fuzzy reasoning \cite{spolaor2020simpful}, but they do not offer an Ex-Fuzzy-style learning pipeline focused on genetic optimisation of compact rule bases and explainability-oriented reporting.

This paper extends Ex-Fuzzy to support Mamdani-style fuzzy regression. The extension introduces three new components: \emph{RuleBaseT1Regression}, \emph{FitRuleBaseRegression}, and \emph{FCMMamdaniRegressor} that mirror the existing classification architecture while adapting gene encoding and inference for scalar consequents. We propose a target-aware partition initialisation strategy using Fuzzy C-Means clustering \cite{bezdek1981} on an augmented input-output space, which places linguistic terms in output-relevant regions of the feature space.

We evaluate the extension on 10 regression datasets from the KEEL repository \cite{keel2011}, comparing two partition strategies (Gaussian and Trapezoidal) against standard baselines (Linear Regression, MLP, Random Forest). The results demonstrate that the FCM-based Gaussian partitions achieve competitive accuracy (mean $R^2 = 0.86$) while producing compact, interpretable rule bases of 10--15 rules.

The remainder of this paper is organised as follows. Section~II provides background on Mamdani fuzzy inference. Section~III describes the software architecture, including partition initialisation and evolutionary optimisation. Section~IV presents experimental results and analysis. Section~V provides usage examples, and Section~VI concludes the paper. Section~VII outlines directions for future work.

\section{Background}

\subsection{Mamdani Fuzzy Rule Inference}

A Mamdani fuzzy rule takes the form:
\begin{equation}
r: \text{IF } x_1 \text{ IS } A_1^{(r)} \text{ AND } \ldots \text{ AND } x_d \text{ IS } A_{D}^{(r)} \text{ THEN } y = z_r
\end{equation}
where $A_k^{(r)}$ is the linguistic term for input $k$ in rule $r$, and $z_r$ is the scalar consequent.

For a given input vector $\mathbf{x}$, each rule's firing strength is computed as the product of antecedent membership degrees \cite{mendel2017}:

\begin{equation}
\tau_r(\mathbf{x}) = \prod_{k \in \boldsymbol{\Omega_r}} \mu_{A_k^r}(x_k)
\end{equation}
where $\boldsymbol{\Omega_r}$ is the set of active antecedents (selected via learning). The final prediction is obtained through weighted average defuzzification \cite{mendel2017}:
\begin{equation}
\hat{y} = \frac{\sum_{r=1}^{R} \tau_r(\mathbf{x}) \cdot z_r}{\sum_{r=1}^{R} \tau_r(\mathbf{x})} 
\end{equation}
where $\hat{y}$ is a crisp numerical output suitable for regression tasks, and $z_r$ is the scalar candidate consequence of each rule. Rules that contain no active antecedent conditions are treated as inactive and assigned zero firing strength to prevent unconditional rule activation.

\subsection{Fuzzy Regression in the literature}
Fuzzy regression has been explored through multiple paradigms, most prominently Takagi--Sugeno--Kang (TSK) models and neuro-fuzzy systems, because their functional consequents often yield strong approximation performance. A systematic review by Chukhrova and Johannssen highlights that a large portion of modern fuzzy regression research concentrates on TSK variants and related hybrid learning procedures, while Mamdani-style regression is comparatively less represented \cite{chukhrova2019}. At the same time, interpretability-driven fuzzy modelling remains an active direction in explainable AI, where rule-based structures are valued for transparency and auditability \cite{mendel2017}. 
Initialisation of fuzzy partitions is commonly performed using clustering or data-driven heuristics, where Fuzzy C-Means is a standard tool for locating representative prototypes and informing fuzzy set placement \cite{bezdek1981}. This motivates the target-aware FCM initialisation adopted in this work.

\subsection{Other Fuzzy Software for Regression}
Several software tools support fuzzy modelling for regression, but with different goals and limitations. MATLAB provides mature fuzzy inference tooling and neuro-fuzzy modelling workflows, but is not native to Python-based ML pipelines \cite{sivanandam2007introduction}. In R, FuzzyR offers an extended fuzzy logic toolbox \cite{chen2020fuzzyr}. For Python, scikit-anfis provides a scikit-learn compatible ANFIS implementation \cite{zhang2024scikit}, and Simpful provides a user-friendly fuzzy reasoning engine supporting Mamdani and Sugeno systems \cite{spolaor2020simpful}. However, these tools do not directly mirror the Ex-Fuzzy focus on data-driven symbolic learning with genetic optimisation of compact rule bases and library-level support for explainability-oriented reporting and benchmarking.

\section{Software Architecture}

This section describes the architectural modifications required to extend Ex-Fuzzy \cite{exfuzzy2024} from classification to regression. The extension preserves compatibility with the existing library structure while introducing new components for continuous output prediction.

\subsection{Gene Encoding for Learning Regression Models}

Ex-Fuzzy's genetic algorithm represents candidate rule bases as integer chromosomes. Table~\ref{tab:gene} compares the gene structure for classification and regression.

\begin{table}[htbp]
\centering
\caption{Gene Structure Comparison}
\label{tab:gene}
\begin{tabular}{@{}lccc@{}}
\toprule
\textbf{Section} & \textbf{Size} & \textbf{Class.} & \textbf{Regr.} \\
\midrule
Feature indices & $A \times R$ & $[0, d{-}1]$ & Same \\
Term indices & $A \times R$ & $[-1, T{-}1]$ & Same \\
Consequents & $R$ & $[0, C{-}1]$ & $[0, 100]$ \\
\bottomrule
\multicolumn{4}{@{}l@{}}{\scriptsize $A$: max antecedents per rule, $R$: rules, $d$: features, $T$: terms, $C$: classes}
\end{tabular}
\end{table}

Here, $T$ denotes the number of linguistic terms per feature (controlled by \texttt{n\_linguistic\_labels}). In the original Ex-Fuzzy classification framework, the chromosome can optionally include genes for fuzzy set optimisation, allowing membership function parameters to be tuned alongside rule structure \cite{exfuzzy2024}. In the present regression extension, the same rule-structure encoding is retained and extended with scalar consequent encoding. Fuzzy partitions are initialised using the proposed strategies and refined through their associated configuration parameters (e.g., global width scaling), while the genetic optimisation focuses on selecting compact, high-performing rule bases.

The key regression modification lies in the consequent section: classification uses integer class indices, whereas regression encodes continuous consequents as integers in the range $[0,100]$ and maps them linearly to the target range $[y_{min}, y_{max}]$. The chromosome layout matches the decision vector of the evolutionary optimizer \texttt{pymoo} \cite{blank2020pymoo}:

\begin{lstlisting}[style=pythonstyle]
# Gene bounds: [features | terms | consequents]
bounds = []
for r in range(nRules):
    for a in range(nAnts):
        bounds.append([0, n_features - 1])  # Feature index
for r in range(nRules):
    for a in range(nAnts):
        bounds.append([-1, n_terms - 1])    # Term index (-1 = inactive)
for r in range(nRules):
    bounds.append([0, 100])                 # Consequent code
\end{lstlisting}

The chromosome for individual $b$ is represented as:

\begin{equation}
\small
\begin{aligned}
\boldsymbol{\rho}^{(b)} =
\Big[
&\underbrace{
i^{(b)}_{1,1},\ldots,i^{(b)}_{1,L},\;
\ldots,\;
i^{(b)}_{R,1},\ldots,i^{(b)}_{R,L}
}_{\text{feature indices}}
,\\
&\underbrace{
t^{(b)}_{1,1},\ldots,t^{(b)}_{1,L},\;
\ldots,\;
t^{(b)}_{R,1},\ldots,t^{(b)}_{R,L}
}_{\text{term indices}}
,\;
\underbrace{
z^{(b)}_{1},\ldots,z^{(b)}_{R}
}_{\text{consequents}}
\Big],
\end{aligned}
\end{equation}
The rule structure is defined by a maximum of $L$ antecedents, indexed by $l \in \{1, \dots, L\}$. Each candidate rule $r=[1,...,R]$ is encoded using three distinct parameters: the input variable index $i^{(b)}_{r,l} \in \{0, \dots, D-1\}$, where $D$ is the number of input factors; the linguistic term index $t^{(b)}_{r,l} \in \{-1, \dots, T-1\}$, where $T$ is the number of terms; and the scalar consequent $z^{(b)}_{r} \in [0, 100]$. A linguistic term value of $t^{(b)}_{r,l} = -1$ explicitly marks the antecedent as inactive. Consequently, the set of active antecedents $\boldsymbol{\Omega_r}$ is defined by filtering the rule components to retain only those with valid linguistic terms, such that $\boldsymbol{\Omega_r} = \{ A_{l}^r \mid t^{(b)}_{r,l} \neq -1 \}$.

\subsection{Inference Mechanism}

Regression uses weighted average defuzzification to produce continuous outputs:

\begin{lstlisting}[style=pythonstyle]
def inference(self, X: np.ndarray) -> np.ndarray:
    firing = self.compute_rule_memberships(X)
    preds = np.zeros(X.shape[0])
    mask = sums > 1e-10
    preds[mask] = (firing[mask] @ self.scalar_consequents) / sums[mask]
    preds[~mask] = self.y_mean
    return preds
\end{lstlisting}

If no rule fires for a given input (i.e., the sum of firing strengths is numerically zero), the prediction defaults to the mean target value observed in the training data. Predicted outputs are clipped to a small margin beyond the training target range to ensure numerical stability.

\subsection{Partition Initialisation}

We evaluate two partition strategies for defining linguistic variables:

\textbf{Gaussian Partitions (Target-Aware FCM):} Fuzzy C-Means clustering is applied to an augmented space $[\mathbf{X}, \alpha \cdot \tilde{y}]$, where $\tilde{y}$ is the standardised target and $\alpha$ (default 2.5) controls the target influence. This target-aware clustering biases partition placement toward output-relevant regions of the input space. Cluster statistics are projected back to input dimensions to construct per-feature Gaussian membership functions, with centres derived from cluster centroids and widths from within-cluster spread. Gaussian widths are further scaled by a global factor ($\texttt{sigma\_scale}$) with mild feature-wise adaptation based on estimated feature relevance, allowing more flexible partitions for informative attributes.

\textbf{Constrained Trapezoidal Partitions:} Trapezoidal membership functions are constructed from feature-wise min/max bounds, with interior terms placed uniformly and overlap controlled by \texttt{trapezoidal\_overlap}. This enforces semantic ordering (Low $<$ Medium $<$ High) but does not adapt to data density.

Both strategies operate on robustly scaled features (median and interquartile range), so linguistic terms represent relative positions within the scaled distribution rather than original units.

\subsection{Evolutionary Rule Learning}

The genetic algorithm optimises rule bases by maximising mean validation of the coefficient of determination on each validation fold. The rule structure (chromosome) remains fixed during fold evaluation; only the validation split changes. Since \texttt{pymoo} minimises objectives, the implementation returns $-F$.

Model complexity is controlled structurally through the maximum number of rules $R$ (\texttt{nRules}) and maximum antecedents per rule $A$ (\texttt{max\_antecedents}). Inactive antecedents ($t=-1$) further reduce effective rule complexity. After evolutionary optimisation, rules with low quality scores are pruned.

\subsection{New Code Components}

The regression extension introduces three new classes that mirror the classification architecture:

\begin{itemize}
    \item \texttt{RuleBaseT1Regression}: Stores rules with scalar consequents and implements weighted average inference with mean-value fallback.
    \item \texttt{FitRuleBaseRegression}: Defines the GA optimisation problem using mean validation $R^2$ as fitness.
    \item \texttt{FCMMamdaniRegressor}: Scikit-learn compatible wrapper providing \texttt{fit()}, \texttt{predict()}, \texttt{score()}, and \texttt{print\_rules()}.
\end{itemize}

\section{Experimental Results and Analysis}

\subsection{Experimental Setup}

We evaluate on 10 regression datasets from the KEEL repository \cite{keel2011} spanning different sizes (34 to 32,614 samples) and dimensionalities (2 to 26 features). Table~\ref{tab:datasets} summarises dataset characteristics. Performance is measured using coefficient of determination ($R^2$) on held-out test sets using official KEEL train-test splits. All experiments use fixed random seeds for reproducibility. Baselines use standard configurations chosen for comparable complexity rather than exhaustive hyperparameter tuning.

\begin{table}[htbp]
\centering
\caption{Dataset Characteristics}
\label{tab:datasets}
\begin{tabular}{@{}lccc@{}}
\toprule
\textbf{Dataset} & \textbf{Train} & \textbf{Test} & \textbf{Features} \\
\midrule
laser & 794 & 199 & 4 \\
anacalt & 3,241 & 811 & 7 \\
friedman & 960 & 240 & 5 \\
autoMPG6 & 313 & 79 & 5 \\
dee & 292 & 73 & 6 \\
diabetes & 34 & 9 & 2 \\
stock & 760 & 190 & 9 \\
mv & 32,614 & 8,154 & 10 \\
compactiv & 6,553 & 1,639 & 21 \\
pole & 11,998 & 3,000 & 26 \\
\bottomrule
\end{tabular}
\end{table}

For fair complexity comparison, we constrain all models similarly: our fuzzy system produces approximately 10--15 rules; MLP uses a single hidden layer of 10 neurons (41--281 parameters); Random Forest uses 10 trees with maximum depth 3 ($\approx$80 leaf nodes); Linear Regression serves as a minimal baseline.

\subsection{Overall Performance Comparison}

Table~\ref{tab:results} presents approximate test $R^2$ scores across all datasets. We report results for the best-performing Gaussian and Trapezoidal configurations, Ex-Fuzzy-G, with specific hyperparameters shown in Table~\ref{tab:configs}.

\begin{table}[htbp]
\centering
\caption{Approximate Test $R^2$ Comparison Across Methods}
\label{tab:results}
\setlength{\tabcolsep}{3pt}
\renewcommand{\arraystretch}{1.05}
\scriptsize
\begin{tabularx}{\columnwidth}{@{}l *{5}{>{\centering\arraybackslash}X} >{\centering\arraybackslash}X@{}}
\toprule
\textbf{Dataset} & \textbf{LR} & \textbf{MLP} & \textbf{RF} & \textbf{ExFG} & \textbf{ExFT} & \textbf{R} \\
\midrule
laser     & 0.75 & \textbf{0.96} & 0.86 & \textbf{0.96} & 0.84 & 14 \\
anacalt   & 0.39 & \textbf{0.96} & \textbf{0.96} & 0.94 & 0.54 & 10 \\
friedman  & 0.68 & \textbf{0.85} & 0.65 & 0.77 & 0.59 & 10 \\
autoMPG6  & 0.78 & 0.83 & 0.82 & \textbf{0.84} & 0.79 & 12 \\
dee       & \textbf{0.84} & 0.83 & 0.80 & 0.82 & 0.77 & 11 \\
diabetes  & 0.19 & $-$0.28 & 0.24 & \textbf{0.53} & 0.31 & 9 \\
stock     & 0.85 & \textbf{0.95} & 0.88 & 0.92 & 0.89 & 15 \\
mv        & 0.82 & \textbf{0.99} & 0.92 & \textbf{0.99} & 0.92 & 10 \\
compactiv & 0.75 & \textbf{0.98} & 0.93 & 0.96 & 0.92 & 14 \\
pole      & 0.49 & \textbf{0.94} & 0.75 & 0.88 & 0.56 & 15 \\
\midrule
\textbf{Mean} & 0.65 & 0.80 & 0.78 & \textbf{0.86} & 0.71 & 12 \\
\bottomrule
\multicolumn{7}{@{}l@{}}{\scriptsize R=Rule count, LR=Linear Regression, RF=Random Forest, ExFG=Ex-Fuzzy Gaussian,}\\
\multicolumn{7}{@{}l@{}}{\scriptsize ExFT=Ex-Fuzzy Trapezoidal. Bold indicates the best (highest) test $R^2$ per dataset}
\end{tabularx}
\end{table}

\begin{table}[htbp]
\centering
\caption{Best Configurations (terms $t$, max rules $r$)}
\label{tab:configs}
\setlength{\tabcolsep}{3pt}
\renewcommand{\arraystretch}{1.05}
\scriptsize
\begin{tabularx}{\columnwidth}{@{}l >{\centering\arraybackslash}X >{\centering\arraybackslash}X >{\centering\arraybackslash}X >{\centering\arraybackslash}X@{}}
\toprule
\textbf{Dataset} & \textbf{Gaus Config} & \textbf{Gaus $R^2$} & \textbf{Trap Config} & \textbf{Trap $R^2$} \\
\midrule
laser     & 5t, 15r & 0.96 & 3t, 15r & 0.84 \\
anacalt   & 5t, 10r & 0.94 & 7t, 15r & 0.54 \\
friedman  & 7t, 10r & 0.77 & 3t, 15r & 0.59 \\
autoMPG6  & 3t, 15r & 0.84 & 3t, 10r & 0.79 \\
dee       & 5t, 15r & 0.82 & 3t, 15r & 0.77 \\
diabetes  & 7t, 10r & 0.53 & 3t, 15r & 0.31 \\
stock     & 5t, 15r & 0.92 & 3t, 15r & 0.89 \\
mv        & 7t, 10r & 0.99 & 3t, 10r & 0.92 \\
compactiv & 7t, 15r & 0.96 & 3t, 20r & 0.92 \\
pole      & 5t, 15r & 0.88 & 3t, 15r & 0.56 \\
\bottomrule
\multicolumn{5}{@{}l@{}}{\scriptsize Gaus=Ex-Fuzzy Gaussian, Trap=Ex-Fuzzy Trapezoidal.}\\
\multicolumn{5}{@{}l@{}}{\scriptsize $R^2$ values rounded to two decimal places.}
\end{tabularx}
\end{table}

The Gaussian fuzzy regression achieves competitive performance while maintaining full interpretability. Key observations:

\textbf{Gaussian vs Baselines:} The Gaussian configuration achieves the best result on 2 datasets (autoMPG6, diabetes) and ties with MLP on laser. On diabetes, where MLP fails ($R^2 \approx -0.28$) due to the small sample size (34 instances), the fuzzy approach succeeds ($R^2 \approx 0.53$).

\textbf{Gaussian vs Trapezoidal:} Gaussian partitions outperform Trapezoidal on all 10 datasets. The gap is largest on anacalt (0.94 vs 0.54), pole (0.88 vs 0.56), and friedman (0.77 vs 0.59). Mean $R^2$ improves from approximately 0.71 to 0.86---a 21\% relative gain.

\textbf{Interpretability:} The fuzzy system uses only 10--15 human-readable rules compared to 41--281 MLP parameters, while achieving comparable mean performance (0.86 vs 0.80).

\subsection{Partition Comparison}

\begin{figure}[htbp]
\centering
\includegraphics[width=\columnwidth]{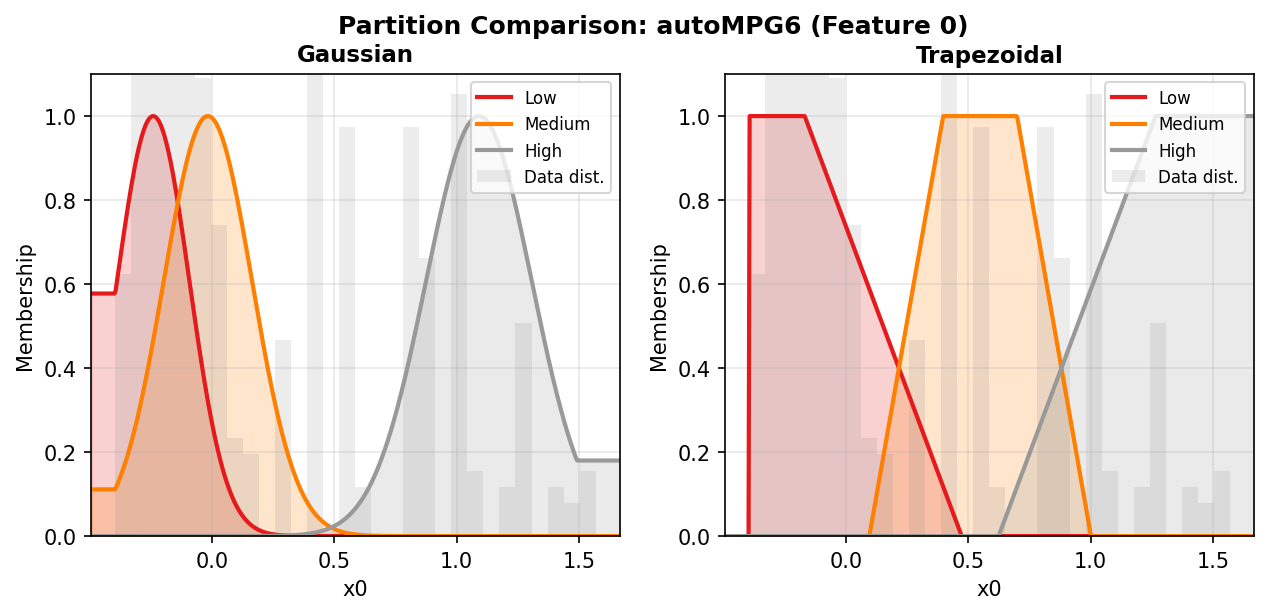}
\caption{Partition comparison for autoMPG6 feature x0 (displacement) in scaled space. Left: Gaussian partitions concentrate Low and Medium terms in the high-density region. Right: Trapezoidal partitions use uniform spacing regardless of data distribution.}
\label{fig:partition_comparison}
\end{figure}

Fig.~\ref{fig:partition_comparison} illustrates the difference between Gaussian and Trapezoidal partitions for feature x0 (displacement) of the autoMPG6 dataset. The grey histogram shows the underlying data distribution in scaled space, which is heavily left-skewed with most samples concentrated in the lower range. The Gaussian partition (left) adapts to this skew: the Low and Medium terms overlap substantially in the high-density region, providing fine-grained discrimination where data concentrates. The High term covers the sparse right tail. In contrast, the Trapezoidal partition (right) distributes terms uniformly across the domain, placing the Medium term's peak in a region where relatively few samples exist.

\begin{figure}[htbp]
\centering
\includegraphics[width=\columnwidth]{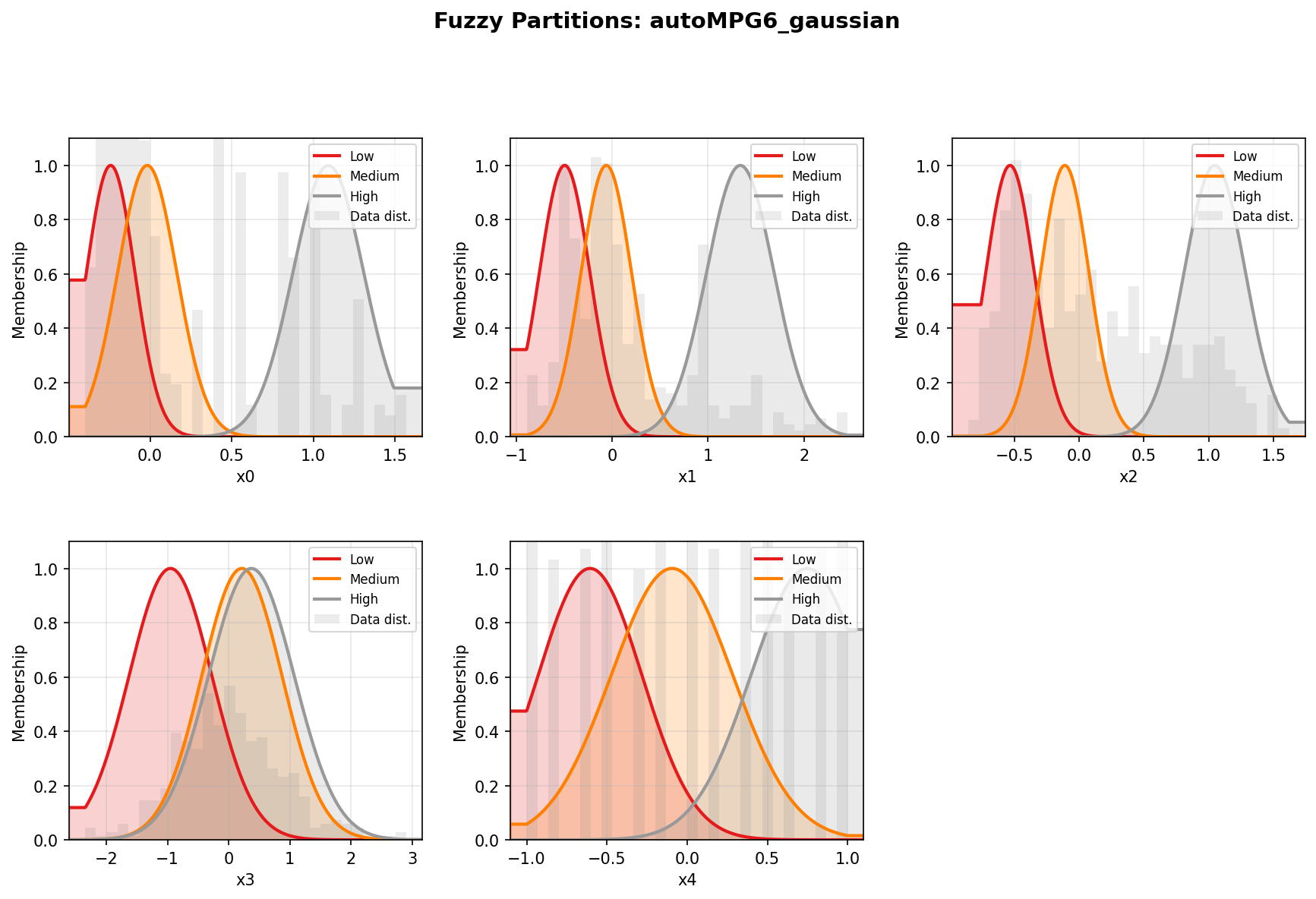}
\caption{Gaussian partitions for all autoMPG6 features in scaled space. Term placement adapts to each feature's distribution: left-skewed features (x0, x1, x2) have terms concentrated on the left; symmetric features (x3) have centred terms.}
\label{fig:partitions_gaussian}
\end{figure}

Fig.~\ref{fig:partitions_gaussian} shows Gaussian partitions for all five autoMPG6 features in scaled space. Several patterns emerge: features x0 (displacement), x1 (horsepower), and x2 (weight) exhibit left-skewed distributions, and the Gaussian terms shift accordingly. Feature x3 (acceleration) shows a more symmetric distribution with correspondingly centred terms. Feature x4 (model year) displays a relatively uniform distribution with slight right-skew, reflected in the term placement.

\begin{figure}[htbp]
\centering
\includegraphics[width=\columnwidth]{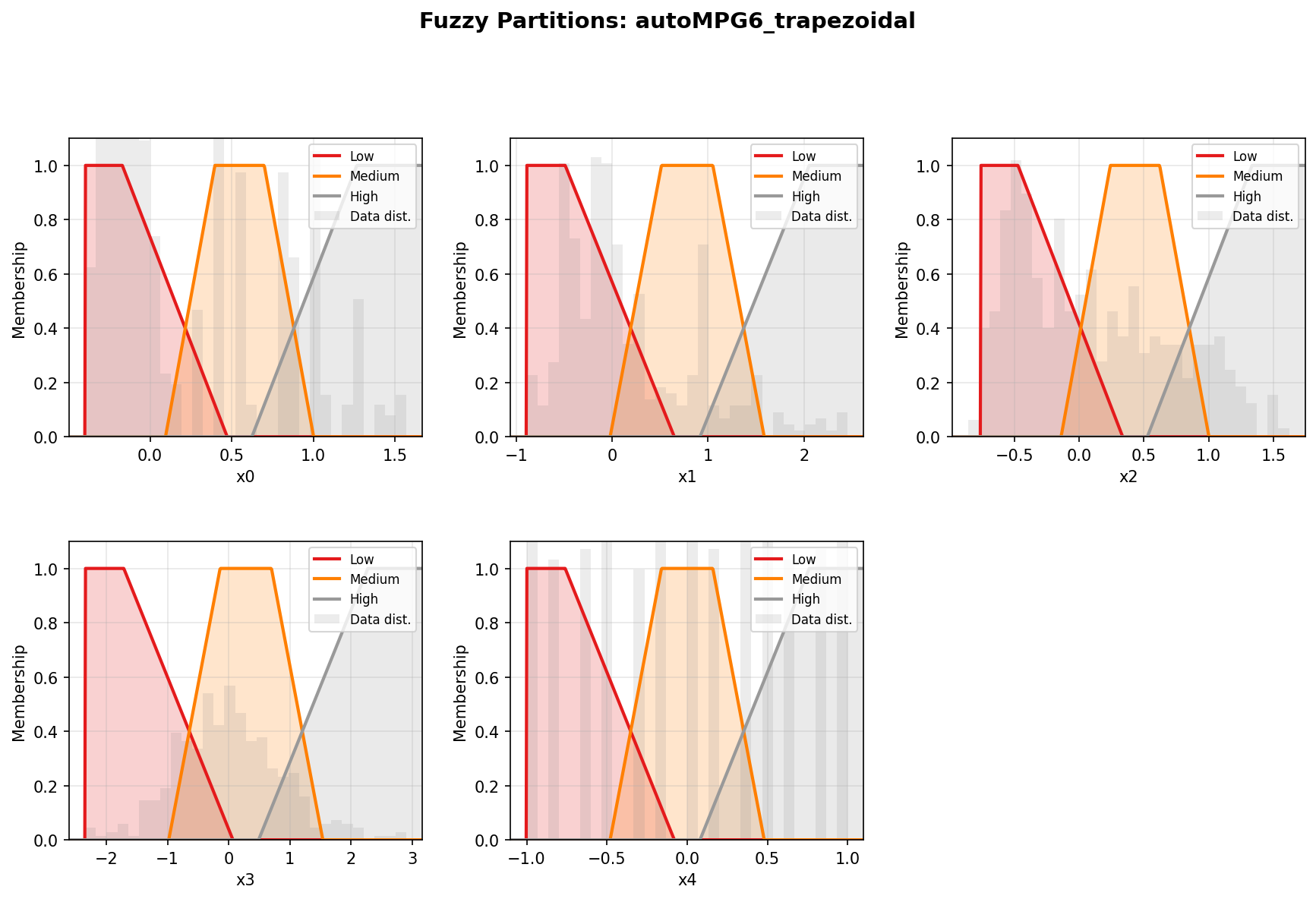}
\caption{Trapezoidal partitions for all autoMPG6 features in scaled space. Uniform spacing ignores data distribution, placing Medium terms in sparse regions.}
\label{fig:partitions_trapezoidal}
\end{figure}

Fig.~\ref{fig:partitions_trapezoidal} shows the corresponding Trapezoidal partitions. The uniform term spacing ignores data density, resulting in the Medium term often covering sparse regions, while the Low term must handle the majority of samples. This mismatch explains the consistent performance gap between Gaussian and Trapezoidal partitions on this dataset.

\subsection{Interpretability Analysis}

Table~\ref{tab:rules_gaussian} presents selected rules from the Gaussian autoMPG6 model, which predicts fuel efficiency (mpg) using 12 rules over 5 input features: displacement (x0), horsepower (x1), weight (x2), acceleration (x3), and model year (x4).

\begin{table}[htbp]
\centering
\caption{Example Rules from autoMPG6 (Gaussian)}
\label{tab:rules_gaussian}
\begin{tabular}{@{}p{5.2cm}c@{}}
\toprule
\textbf{Rule Antecedent} & \textbf{mpg} \\
\midrule
IF x0 IS High AND x1 IS High AND x2 IS High AND x4 IS Low & 11.3 \\
IF x0 IS High AND x1 IS High AND x2 IS High AND x4 IS Med & 16.5 \\
IF x0 IS Low AND x2 IS Low AND x4 IS High & 37.2 \\
IF x0 IS Med AND x2 IS Low AND x4 IS High & 35.3 \\
IF x1 IS Med AND x2 IS Med AND x4 IS Med & 19.5 \\
\bottomrule
\multicolumn{2}{@{}l@{}}{\scriptsize x0=displacement, x1=horsepower, x2=weight, x4=year.} \\
\multicolumn{2}{@{}l@{}}{\scriptsize Consequent values approximate. Terms defined in scaled space.}
\end{tabular}
\end{table}

The rules capture intuitive domain relationships: older vehicles (x4 IS Low) with high displacement, horsepower, and weight achieve low fuel efficiency ($\approx$11.3 mpg), while newer lightweight vehicles (x4 IS High, x0 IS Low, x2 IS Low) achieve high efficiency ($\approx$37.2 mpg). Notably, acceleration (x3) appears in only one Gaussian rule, suggesting limited predictive value for this task. Note that linguistic terms such as ``High'' and ``Low'' are defined relative to the scaled feature distributions, not original units.

For comparison, Table~\ref{tab:rules_trapezoidal} shows selected rules from the Trapezoidal model. While also interpretable, these rules achieve lower accuracy ($R^2 \approx 0.79$ vs 0.84) due to suboptimal partition placement.

\begin{table}[htbp]
\centering
\caption{Example Rules from autoMPG6 (Trapezoidal)}
\label{tab:rules_trapezoidal}
\begin{tabular}{@{}p{5.2cm}c@{}}
\toprule
\textbf{Rule Antecedent} & \textbf{mpg} \\
\midrule
IF x0 IS High AND x1 IS Med & 13.5 \\
IF x0 IS High AND x3 IS Low & 13.1 \\
IF x0 IS Low AND x1 IS Low AND x2 IS Low AND x4 IS High & 42.5 \\
IF x0 IS Med AND x1 IS Low AND x2 IS Low AND x4 IS High & 34.9 \\
IF x1 IS Med AND x2 IS Med & 18.4 \\
\bottomrule
\multicolumn{2}{@{}l@{}}{\scriptsize x0=displacement, x1=horsepower, x2=weight, x3=accel, x4=year.} \\
\multicolumn{2}{@{}l@{}}{\scriptsize Consequent values approximate. Terms defined in scaled space.}
\end{tabular}
\end{table}

\section{Usage Example}

The regression extension follows scikit-learn conventions.

The following example demonstrates training and evaluating the proposed regressor using a standard scikit-learn workflow. The model is fitted on training data and evaluated using built-in \texttt{score} method.

\begin{lstlisting}[style=pythonstyle]
import numpy as np
from sklearn.model_selection import train_test_split
from sklearn.datasets import load_diabetes
from fcm5 import FCMMamdaniRegressor

# Load a regression dataset
X, y = load_diabetes(return_X_y=True)
X_train, X_test, y_train, y_test = train_test_split(
    X, y, test_size=0.2, random_state=42)

# Train model with Gaussian partitions
model = FCMMamdaniRegressor(
    nRules=15,
    n_linguistic_variables=3,
    max_antecedents=4,
    partition_type='gaussian',
    sigma_scale=1.5,
    verbose=True)
model.fit(X_train, y_train)

# Evaluate
print(f"Test R2: {model.score(X_test, y_test):.3f}")

# Predict
preds = model.predict(X_test)
\end{lstlisting}

After training, rules can be printed in human-readable form to support inspection and domain validation, and the underlying rule base object can be accessed programmatically. 

\begin{lstlisting}[style=pythonstyle]
# Print human-readable rules
model.print_rules()

# Access rule base object
rb = model.get_rulebase()
print(f"Number of rules: {len(rb)}")
\end{lstlisting}

Partition strategies can be compared by training identical models that differ only in the membership function construction, allowing evaluation of data-adaptive Gaussian partitions against uniform trapezoidal partitions.

\begin{lstlisting}[style=pythonstyle]
# Gaussian partitions (target-aware)
model_gaus = FCMMamdaniRegressor(
    nRules=15,
    n_linguistic_variables=3,
    partition_type='gaussian')
model_gaus.fit(X_train, y_train)

# Trapezoidal partitions (uniform)
model_trap = FCMMamdaniRegressor(
    nRules=15,
    n_linguistic_variables=3,
    partition_type='trapezoidal',
    trapezoidal_overlap=0.5)
model_trap.fit(X_train, y_train)

print(f"Gaussian R2: {model_gaus.score(X_test, y_test):.3f}")
print(f"Trapezoidal R2: {model_trap.score(X_test, y_test):.3f}")
\end{lstlisting}

\section{Conclusion}

This paper presented a regression extension for the Ex-Fuzzy library, enabling Mamdani-style fuzzy inference for continuous output prediction. The key contributions are threefold: (1) a software architecture that adapts Ex-Fuzzy's genetic algorithm framework from classification to regression through modified gene encoding and weighted average defuzzification; (2) a target-aware partition initialisation strategy using Fuzzy C-Means clustering on an augmented input-output space; and (3) comprehensive experimental validation across 10 benchmark datasets. The extension follows scikit-learn conventions, providing a familiar API for practitioners. All components integrate seamlessly with the existing Ex-Fuzzy infrastructure, requiring modifications only to consequent handling and inference mechanisms.

The experimental results demonstrate that the Gaussian (FCM-derived) partition strategy consistently outperforms traditional uniform trapezoidal partitions, achieving a mean $R^2$ of 0.86 compared to 0.71---a 21\% relative improvement. The proposed method achieves competitive accuracy with neural networks (mean $R^2$ 0.86 vs 0.80 for MLP) while maintaining high interpretability through 10--15 human-readable IF-THEN rules.

There are several open research directions from both methodological and software perspectives. Methodologically, future work will explore extensions to Type-2 fuzzy systems to improve uncertainty handling \cite{fumanal2025reliable}, investigate alternative data-driven partition initialisation strategies, such as subtractive clustering, and study uncertainty quantification using calibration metrics. From a software perspective, the proposed regression extension will be contributed to the official Ex-Fuzzy repository to support reuse and further development by the fuzzy systems community.

These directions connect to a broader line of fuzzy systems research. The planned Type-2 extension for uncertainty handling relates directly to temporal type-2 fuzzy systems developed for time-dependent explainable AI~\cite{kiani2022temporal}, which could inform how membership functions are extended along the time axis if the regression setting is extended to streaming or sequential data. Such a streaming extension would itself parallel evolving, self-adaptive fuzzy classifiers that relearn online from streaming sensor data without offline retraining~\cite{andreu2010real, andreu2013evolving}, offering a template for an online variant of the current batch, genetic-algorithm-based training procedure. The human-readable IF-THEN rules produced here are also a natural fit for computing-with-words formalisms for defining and combining linguistic variables~\cite{gupta2022gentle}. Because interpretability is motivated here by safety-critical and regulated domains, validating the regression extension on more realistic, harder benchmark data, beyond the curated KEEL datasets, would strengthen this claim, echoing a similar lesson from clinical brain-computer interface research, where competitions moved from healthy-subject to more challenging patient data to properly stress-test reliability~\cite{chowdhury2021clinical}; the same broader concern, that models must remain reliable under noisy, real-world conditions, extends beyond fuzzy regression to deep learning systems generally~\cite{chaudhary2023review}. Finally, packaging this extension for reuse, in the spirit of accessible, GUI-driven toolboxes built for other fuzzy, multivariate analysis pipelines~\cite{andreu2016ealab}, would help lower the barrier to adoption alongside the planned contribution to the Ex-Fuzzy repository.

\section{Acknowledgement}

This research and Javier Fumanal-Idocin were supported by EU Horizon Europe under the Marie Skłodowska-Curie COFUND grant No 101081327 YUFE4Postdocs. This research is kindly supported by UKRI BBSRC project EyeWarn (code: APP37953).

\bibliographystyle{IEEEtran}
\bibliography{conference_101719}

\end{document}